\documentclass[sigconf]{acmart}
\usepackage{graphicx}
\usepackage{balance}  
\usepackage{soul, color,wrapfig}
\setstcolor{red}

\usepackage{dsfont}
\usepackage{bbm}
\usepackage{booktabs} 
\usepackage{caption}
\usepackage{amsmath}
\usepackage{mathtools}
\usepackage{algorithm}
\usepackage{sidecap}
\usepackage{algpseudocode}
\usepackage{multirow}
\usepackage{epsfig,graphics,float}
\usepackage{subfig,balance,lineno}
\usepackage{float}

\newlength\myindent
\newcommand{\multiline}[1]{%
  \begin{tabularx}{\dimexpr\linewidth-\ALG@thistlm}[t]{@{}X@{}}
    #1
  \end{tabularx}
}
 
\newtheorem{definition}{Definition}[section]
\newtheorem{example}{Example}[section]

\AtBeginDocument{%
  \providecommand\BibTeX{{%
    \normalfont B\kern-0.5em{\scshape i\kern-0.25em b}\kern-0.8em\TeX}}}

\setcopyright{acmcopyright}
\copyrightyear{2018}
\acmYear{2018}
\acmDOI{10.1145/1122445.1122456}

\acmConference[Woodstock '18]{Woodstock '18: ACM Symposium on Neural
  Gaze Detection}{June 03--05, 2018}{Woodstock, NY}
\acmBooktitle{Woodstock '18: ACM Symposium on Neural Gaze Detection,
  June 03--05, 2018, Woodstock, NY}
\acmPrice{15.00}
\acmISBN{978-1-4503-XXXX-X/18/06}

\begin{document}

\title{Efficient Knowledge Graph Validation via Cross-Graph Representation Learning}

\author{Yaqing Wang$^{1}$, Fenglong Ma$^2$ and Jing Gao$^1$}
\affiliation{\institution{\textsuperscript{1}State University of New York at Buffalo, New York, USA \\
\textsuperscript{2}Pennsylvania State University, Pennsylvania, USA \\
\textsuperscript{1}\{yaqingwa,  jing\}@buffalo.edu, 
\textsuperscript{2} fenglong@psu.edu
}
}

\begin{abstract}
  Recent advances in information extraction have motivated the automatic construction of huge Knowledge  Graphs  (KGs) by mining from large-scale text corpus. However, noisy facts are unavoidably introduced into KGs that could be caused by automatic extraction. 
To validate the correctness of facts (i.e., triplets) inside a KG, one possible approach is to map the triplets into vector representations by capturing the semantic meanings of facts. Although many representation learning approaches have been developed for knowledge graphs, these methods are not effective for validation. They usually assume that facts are correct, and thus may overfit noisy facts and fail to detect such facts. 

Towards effective KG validation, we propose to leverage an external human-curated KG as auxiliary information source to help detect the errors in a target KG. The external KG is built upon human-curated knowledge repositories and tends to have high precision. On the other hand, although the target KG built by information extraction from texts has low precision, it can cover new or domain-specific facts that are not in any human-curated repositories.  To leverage external KG for validation, one intuitive approach is to find matched triplets between the target KG and the external KG. However, this approach can only apply to the small portion of triplets that are covered by both external and target KGs and is not useful for the validation of majority of the triplets. To tackle this challenging task, we propose a cross-graph representation learning framework, i.e., CrossVal, which can leverage an external KG to validate the facts in the target KG efficiently. This is achieved by embedding triplets based on their semantic meanings, drawing cross-KG negative samples and estimating a confidence score for each triplet based on its degree of correctness.
We evaluate the proposed framework on datasets across different domains. Experimental results show that the proposed framework achieves the best performance compared with the state-of-the-art methods on large-scale KGs. 
\end{abstract}




\maketitle

\section{Introduction}

Recent advances in information extraction have motivated the automatic construction of huge Knowledge Graphs (KGs) from large-scale text corpus. Examples include KnlowledgeVault~\cite{dong2014knowledge}, NELL~\cite{mitchell2018never} and commercial projects~\cite{dong2018challenges,dong2020autoknow, gao2018building}. These KGs contain
millions or billions of relational facts in the form of \emph{subject-predicate-object}
(SPO) triplets, e.g., (\emph{albert\_einstein, marriedTo, mileva\_marici}), and thus they are the backbone for many downstream applications, including question answering~\cite{zheng2018question} and recommendation~\cite{ noia2016sprank}. However, there often exist noisy facts in those KGs, which are introduced by the information extraction methods~\cite{stepanova2018rule}.  For example, the precision of NELL is estimated to be only around 50\%  on some relations ~\cite{mitchell2018never}. The existence of such errors in KGs may significantly degrade the performance of downstream applications. 
However, the magnitude of KGs does not allow for manual validation. Thus,  there is an urgent need for the development of automatic yet effective knowledge graph validation algorithms.

Some automatic approaches~\cite{ortona2018rudik, tanon2019learning, paulheim2017knowledge, pujara2013knowledge} are proposed to validate the KGs by utilizing type constraints or mined rules. 
Although these approaches can be used to validate a subset of triplets covered by type constraints or rules,  there may exist many facts that are not covered by type constraints or rules but need validation. 
An intuitive way to validate all the facts is to learn a function that maps each triplet, including the subject, relation, and object, to a low-dimensional feature vector space, which represents the semantic meaning of the given triplet. Using the learned function, the questionable facts can be inferred when the triplets do not satisfy the relationship in the new representation space.
We may adopt widely used knowledge graph embedding (i.e., representation learning) technique ~\cite{nickel2016review, wang2017knowledge} as the function to infer such representations that capture the semantic meanings of triplets. However, such methods are not suitable for the validation task. There is a strong assumption held by existing knowledge graph embedding methods, i.e., the observed triplets are all correct~\cite{bordes2013translating, yang2014embedding, balavzevic2019tucker}. Therefore, there is a high chance of overfitting if these approaches are directly applied to knowledge graphs that contain noisy facts. To address this issue, Xie et al.~\cite{xie2018does} propose a confidence-aware knowledge graph learning method, which estimates the confidence of observed triplets based on internal structure information and further detects errors in the KG. However, as discussed in~\cite{xie2018does}, it is very challenging to spot the errors without external validation information. 


To conduct effective validation on the target KG, one valuable external source we can rely on is other human-curated knowledge graphs, referred to as external KGs, whose facts are more likely to be correct. Such KGs~\cite{yago, lehmann2015dbpedia} are typically built by extracting from the semi-structured content of Wikipedia pages. As human curation is involved in Wikipedia and the extraction from semi-structured content incurs less errors, these KGs usually have high precision. However, compared with KGs constructed from texts, in these KGs, it is not easy to include new knowledge or knowledge specific to a certain domain (such as medical domain) that are not in Wikipedia or other human-curated knowledge repositories. Leveraging these high-precision KGs for the validation of the target KG built upon text corpus is thus very helpful. In this way, we can improve the precision of a KG that is built by information extraction from texts, and thus can benefit many applications that need domain-specific or up-to-date knowledge that is only contained in texts. One intuitive approach~\cite{liu2017measuring} is to find matched triplets between the target KG and the external KGs, and correct the errors in these matched triplets. However, matched triplets are usually only a small portion of  the triplets and thus this method is not useful for the validation of majority of the triplets. 


In light of the aforementioned challenges, we propose a \underline{cross}-graph \underline{val}idation framework~(\textsf{CrossVal}), which can leverage an external KG for the task of automatic knowledge graph validation. Specifically, to facilitate the information transfer between the target KG and the external KG for validation, we propose a novel negative sampling method, which generates the negative triplets that consist of relations and entities from different KGs. More specifically, we generate informative negative triplets based on the mutually conflicting relations that exist in both KGs. The questionable facts in the target KG can be spotted by comparison with generated negative samples. By applying the proposed negative sampling method, an effective bridge over two graphs can be built.  
In this way, we can not only validate matched triplets, but also borrow the information from the external KG to validate the majority of triplets that are not covered by the external KG.  Moreover, many state-of-the-art KG embedding models can be plugged into the proposed framework for validation. We relax the assumption held by KG embedding methods and incorporate the confidence estimation component into model training to 
alleviate the effect of misinformation during embedding.

In the end, we validate the proposed framework on KGs in both \emph{medical} and \emph{general} domains to show its applicability and effectiveness. More specifically, we evaluate our proposed framework on synthetic datasets, i.e., real KGs with synthetic errors, following the similar setting in the previous work~\cite{xie2018does, melo2017detection}. Experimental results show that the proposed framework outperforms the state-of-the-art approaches.
 We also introduce a manually labeled dataset based on NELL KG and conduct extensive experiments on this real dataset. Experimental results show that: (1) Regarding error detection by all the methods, there exists a big gap in the performance on synthetic and real datasets. This big gap demonstrates the significant difference between synthetic and real errors. (2) Our proposed framework can generally improve the validation performance  around 8\% on the \textit{Recall of Ranking metric} and over 20\% on the \textit{Filtered Mean Rank metric} compared with the state-of-the-art KG embedding methods on the real dataset. (3) Our proposed framework can conduct effective validation even with a small portion of external information. (4) The efficiency experiments  demonstrate that our proposed framework can scale well to large-scale KGs. Finally, quantitative and qualitative experiments are conducted to further demonstrate the effectiveness of the proposed framework. We will publicly release the manually labeled dataset to the community to encourage further research on knowledge graph validation.

\vspace{-0.0in}
\section{Related Work}
\label{sec:related_work}

In this section, we briefly review the work related to the proposed model.

The errors in the knowledge graph can negatively impact the related applications and knowledge acquisition~\cite{manago1987noise}. To tackle this problem, knowledge graph validation and other related tasks (e.g., error detection) recently attract wide attention~\cite{xie2018does, wang2020automatic, melo2017detection, tanon2019learning, wang2017discovering, paulheim2017knowledge, chu2016data}.  Since the magnitude of current knowledge graphs does not allow for human curation, lots of researchers focus on automatic knowledge graph validation~\cite{nickel2016review, paulheim2017knowledge, tanon2019learning}. More specifically, related work includes numerical error detection~\cite{li2015probabilistic}, validation via triple query from external KGs~\cite{liu2017measuring}, fact validation via web-search~\cite{gerber2015defacto}, etc. However, most of the existing methods~\cite{lehmann2010ore, ma2014learning}  do not scale well on large-scale KGs and may only cover triplets that have certain patterns.  In ~\cite{ortona2018rudik}, the authors propose to detect errors based on the discovery of declarative rules over knowledge-bases. Though such a method can precisely detect errors, it can only validate a subset of triplets, which are covered by discovered rules.  Melo et al.~\cite{melo2017detection} propose to detect error based on the hand-crafted path and type features. Although this approach can cover all the triplets, it can only detect errors that are captured by the hand-crafted features and complicated errors may escape the detection.  


\begin{table}
\caption{Scoring functions of several state-of-the-art KG representation learning methods.  $d_e$ is the dimensionality of entity embeddings. $\overline{\textbf{e}}_o \in \mathbb{C}^{d_e}$ 
is the complex conjugate of $\textbf{e}_o$ and $\operatorname{Re}(\cdot)$ means taking the real of a complex value. $h_{e_s}$, $t_{e_s}$ are the head and tail entity embedding of entity $e_s$.  $M_r \in \mathbb{R}^{d_e \times d_e}$ is a linear mapping associated with the relation $r$. $\textbf{r}^{-1}$ represents the embedding relation $r^{-1}$ which is inverse relation of $r$. $\langle \cdot \rangle$ denotes the dot product operation. }
\vspace{-0.1in}
\label{tab: embedding}
\centering
\resizebox{\linewidth}{!}{
\begin{tabular}{c|c|c|c}
\toprule

Method& Score Function & Relation Parameters &Type \\

\midrule
\midrule
TransE~\cite{bordes2013translating} & $-|| \textbf{e}_s + \textbf{r} - \textbf{e}_o||_{1/2}$& $\textbf{r} \in \mathbb{R}^{d_e}$  & Translational \\

DistMult~\cite{yang2014embedding} &$\langle \textbf{e}_s,\textbf{r},\textbf{e}_o \rangle$&$\textbf{r} \in \mathbb{R}^{d_e}$ & Multiplicative \\
ComplEx~\cite{trouillon2017knowledge} &$\operatorname{Re}(\langle \textbf{e}_s,\textbf{r},\overline{\textbf{e}}_o \rangle)$&$\textbf{r} \in \mathbb{C}^{d_e}$ & Multiplicative \\
Analogy~\cite{liu2017analogical} &$\textbf{e}_s^T M_r \textbf{e}_s$&$M_r \in \mathbb{R}^{d_e \times d_e}$ & Multiplicative \\
SimplE~\cite{kazemi2018simple} &$\frac{1}{2}(\langle \textbf{h}_{e_s},\textbf{r},\textbf{t}_{e_o} \rangle + \langle \textbf{h}_{e_o},\textbf{r}^{-1},\textbf{t}_{e_s} \rangle))$ &$\textbf{r} \in \mathbb{R}^{d_e}$& Multiplicative \\
\bottomrule
\end{tabular}
}
\vspace{-0.25in}
\end{table}

Knowledge graph representation learning, which aims to learn the semantic meaning of entities and relations, have received lots of attention recently~\cite{nickel2016review, wang2017knowledge}. The knowledge graph embedding methods can be broadly summarized into two types based on the type of score function they adopt: translational and multiplicative approaches. The translational approaches~\cite{ lin2015learning, ji2015knowledge} define additive score functions over embeddings and the multiplicative approaches~\cite{yang2014embedding, trouillon2017knowledge, liu2017analogical, kazemi2018simple} define product-based score functions over embeddings.  
We list the score function, relation parameters and corresponding  types of several state-of-the-art knowledge graph embedding methods in Table~\ref{tab: embedding}.  Due to strong assumption on the correctness of observed triplets, most of the existing graph embedding frameworks cannot be directly applied to knowledge graph validation task. To tackle this limitation, the confidence aware knowledge graph embedding methods~\cite{xie2018does, zhao2019scef} are proposed to detect errors in the knowledge graph. Different from existing work, we leverage information from external KGs for large scale KG validation purpose. To leverage external KG for validation, we propose a novel cross-graph representation learning framework which can effectively coordinate with state-of-the-art multiplicative KG embedding methods.

\section{Methodology}\label{sec:methodology}
In this section, we first define knowledge graph validation problem and then introduce how to leverage an auxiliary knowledge graph to validate the facts in the target KG.
\subsection{Overview}
\textbf{Knowledge Graph Validation.} The problem addressed in this paper is to validate knowledge triplets contained in a target knowledge graph. Following the previous
work~\cite{paulheim2017knowledge}, we define the knowledge graph validation as a \emph{ranking problem}, i.e., given a knowledge graph which may contain errors, the goal is to rank all the knowledge triplets according to their probabilities of being true.

\begin{figure}[hbt]
\includegraphics[width=2.9in]{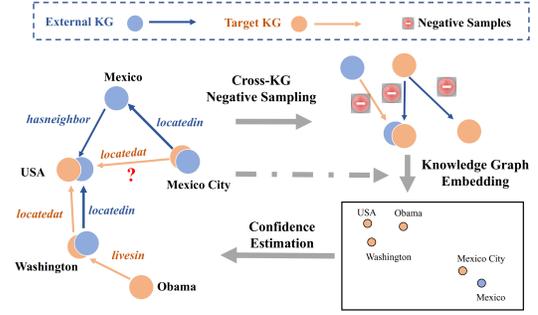}
\centering
\vspace{-0.1in}
\caption{The proposed framework \textsf{CrossVal}. Nodes represent entities, and edges represent existing relationships. }
\label{fig:framework}
\end{figure}

Fig.~\ref{fig:framework} shows the proposed knowledge graph validation framework \textsf{CrossVal}, which leverages an external KG to estimate the probabilities of being true for all the triplets in the target KG.
There are three main components in the proposed framework: \emph{knowledge graph embedding}, \emph{cross-KG negative sampling}, and \emph{confidence estimation}.
The knowledge graph embedding framework can learn the semantic meanings of both entities and relations, which are valuable for error detection. To facilitate information transfer between the external KG and the target KG for validation, 
we propose a cross-KG negative sampling method, which can be considered as an effective bridge to connect both graphs. Existing knowledge graph embedding approaches~\cite{yang2014embedding, trouillon2017knowledge, liu2017analogical} randomly generate negative samples. Instead, we propose to use the mutually conflicting relations existing in both graphs to create negative samples.
To validate the correctness of triplets, we further design a confidence estimation component, which aims to degrade the influence of misinformation by automatically learning an score for each triplet.
The proposed framework tightly integrates these three components that work jointly to enhance the validation performance. 
In the following subsections, we first present the conventional knowledge graph embedding framework and its limitations for the validation task. Then we illustrate how to address these limitations and introduce the proposed cross-graph validation framework. 

\subsection{Knowledge Graph Embedding}
\label{sec:kge}
Knowledge Graph embedding methods aim to project the entities and relations in a KG to low-dimensional dense vectors. To mathematically introduce the KG embedding methods, we represent the entities of the KG as $\mathcal{E}$ and  the relations as $\mathcal{R}$. The KG can be represented as a set of triplets $(e_s, r, e_o)$  denoted as $S$, where $e_s, e_o \in \mathcal{E}$ denote subject and object entities respectively, and $r \in \mathcal{R}$ denotes the relation between them. For every triplet $ (e_s, r, e_o)$, the function $\phi$ of KG embedding models assigns a score, $\phi(e_s, r, e_o) \in \mathbb{R}$, indicating whether this triplet is true or not.  


Most of knowledge graph embedding works~\cite{transH, yang2014embedding, bordes2013translating} follow the open world assumption (OWA), which states that KGs contain only true facts, and non-observed facts can be either false or just missing. To learn the embeddings of the given KG, it needs to generate negative samples based on the \textit{local-closed world assumption}~\cite{nickel2016review}, i.e., a KG is assumed to be locally complete. More specifically, if $(e_s, r, e_o)$ is a triplet in the given KG, the subject or object entity can be replaced by a random entity. In such a way, the negative triplet $(\cdot, r, e_o)$ or $(e_s, r, \cdot)$ is generated. If the generated negative samples do not exist in the original KG, they can be used as negative samples. The set of negative samples is denoted as $S^\prime$.  

The objective function of knowledge graph embedding models can be formulated as a pairwise ranking loss:

\begin{equation*}
    L =\sum_{s \in S} \sum_{s^\prime \in S^\prime}   \max(\phi(s^\prime) -\phi(s) + \gamma,0),
\end{equation*}
 which intends to minimize the  confidence value margin between negative samples and positive samples. When the confidence values of negative values are lower than those of positive samples, the loss values are zero.  Here $\gamma$ denotes the margin. The objective function can also be defined as the minimization of the logistic loss between the embedding and the label of a triplet:
 $$
 L =\sum_{s  \in S \bigcup S^\prime}   (1 + exp(-y_{s} \cdot \phi(s^\prime))),
 $$
 where $y_s = \pm 1$ is the label (i.e., positive or negative) of triplet $s$.  
 
 The assumption held by most KG embedding studies is that \emph{all of the observed triplets $S$ are correct}. However, this assumption may not hold  in the real world because \textbf{no KG is really free from errors}. Therefore, to validate the KG, we propose a new framework, which takes the probability estimation of triplets into consideration. 
Besides, to obtain more information for the validation, an external KG is incorporated. 
The challenge of using another KG for validation is how to construct a bridge between the target and the external KG. Thus, we introduce our entity alignment procedure and cross-KG negative sampling method.

\vspace{-0.05in}
\subsection{Entity Alignment}

The goal of entity alignment is to discover overlapping entities that represent the same real-world entity in two knowledge graphs. Through this procedure, the bridge between two different KGs can be constructed. Before introducing entity alignment procedure, we first introduce notations and define the overlapping entity.

\textbf{Notations.} Let $G_1$ and $G_2$ represent two KGs. $G_1$ is the target graph for validation, and $G_2$ is the external KG that is used to provide validation information. The entity sets of two KGs are denoted as $\mathcal{E}^1$ and $\mathcal{E}^2$, and the relation sets are $\mathcal{R}^1$ and $\mathcal{R}^2$, respectively.

\begin{example}
Take $G_1$ as an example. In Fig.~\ref{fig:framework}, $\mathcal{E}^1$ = $\{$USA, Washington, Obama, Mexico City$\}$ and $\mathcal{R}^1$ = $\{$locatedat, livesin$\}$. Three observed triplets are (Mexico City, locatedat, USA), (Washington, locatedat, USA) and (Obama, livesin, Washington).
\end{example}

\begin{definition} [\textbf{Overlapping Entity}]
Given two entities $e^1 \in \mathcal{E}^1$ and $e^2 \in \mathcal{E}^2$, entities $e^1$ and $e^2$ are  \emph{overlapping entities} if $e^1$ and $e^2$ represent the same real-world entity. The relationship of entities $e^1$ and $e^2$ can be denoted as $e^1 = e^2$.
\end{definition}

 To align the overlapping entities in two KGs, we can take advantage of their name conventions. For example, in the medical KGs, they usually employ the Concept Unique Identifier (CUI) to represent the entities with the same meaning. In the general domain, the entities are represented as string expressions. We employ exact string matching algorithm on entity name conventions to align the overlapping entities existing in  different KGs. Then we further take advantage of KG ontologies, which include the relations like \emph{acronyms} and \emph{known aliases}. Such relations usually indicate the same entities with different string expressions. Once overlapping entities are identified, we ensure that these aligned entities share the same embeddings because they refer to the same entities in the real world. 

\subsection{Cross-KG Negative Sampling}\label{subsec:ckg}
In this subsection, we introduce the proposed  cross-KG negative sampling method.
This method not only constructs an effective bridge between different KGs efficiently but also generates more informative negative samples compared with existing knowledge graph embedding models. Our proposed  method is designed based on the negative relationship between relations from different KGs. The negative relationship indicates that two relations do not have any \emph{overlapping entity pair}. The definitions of \emph{overlapping entity pair} is provided as follows. 

\vspace{-0.05in}
\begin{definition} [\textbf{Overlapping Entity Pair}]
Given two triplets $(e_s^1, r^1, e_o^1)$ and $(e_s^2, r^2, e_o^2)$ from different KGs, if $e_s^1 = e_s^2$ and $e_o^1 = e_o^2$, we define $(e_s^1, e_o^1)$ and $(e_s^2, e_o^2)$ as  \emph{overlapping entity pair} for $r^1$ and $r^2$. The set of entity pair for $r^1$ and $r^2$ can be written as $O(r^1, r^2)$.
\end{definition}
\vspace{-0.1in}
\begin{example}\label{example:oec}
In Fig.~\ref{fig:framework}, (Washington, locatedat, USA) is in the target graph, and (Washington, locatedin, USA) is in the external graph. For the relations ``locatedat'' and ``locatedin'', they share the overlapping entities ``Washiongton'' and ``USA''. Thus, the overlapping entity pair of relations ``locatedat'' and ``locatedin'' is  $\{(Washington, USA)\}$, i.e., $O(locatedat, locatedin)$ = \{(Washington, USA)\}.
\end{example}

\vspace{-0.05in}
According to the definition of overlapping entity pair, we can introduce the concept of cross-KG negative relation.
\vspace{-0.05in}
\begin{definition}[\textbf{Cross-KG Negative Relation}]
For two relations $r^1 \in \mathcal{R}^1, r^2 \in \mathcal{R}^2$, if they do not have any overlapping entity pair, i.e., $O(r^1, r^2) = \emptyset$, then the relations $r^1, r^2$ can be written as $r^1 \perp r^2$, which is denoted as cross-KG negative relation for each other.  
\end{definition}
\vspace{-0.05in}

For a given relation $r_*^1 \in \mathcal{E}^1$, the cross-KG negative relation set $N(r_*^1)$ of $r_*^1$ can be represented as 
$$
N(r_*^1) = \{r^2 | r^2 \perp r_*^1, r^2 \in \mathcal{E}^2\},
$$
and the cross-KG negative set $N(r_*^2)$  of the relation $r_*^2 \in \mathcal{E}^2$ can be represented as 
$$
N(r_*^2) = \{r^1 | r^1 \perp r_*^2, r^1 \in \mathcal{E}^1\}.
$$

\begin{example}\label{example:negative_set}
Take the relation ``livesin'' in Fig.~\ref{fig:framework} as an example. The pair of entities on this relation is (Obama, Washington). This pair of entities does not satisfy any relation in the external graph. Thus, all the relations in the external graph are its cross-KG negative relations, i.e., $N(livesin) = \{locatedin, hasneighbor\}$. For the relation ``hasneighbor'' in the external graph, its cross-KG negative relation set is $N(livesin, locatedat)$.
\end{example}

Next, we present how to generate cross-KG negative samples over two KGs based on cross-KG negative relation sets.  The cross-KG negative samples can be generated based on two strategies: \emph{relation replacement} and \emph{entity replacement}. 

\begin{figure}[hbt]
\includegraphics[width=2.5in]{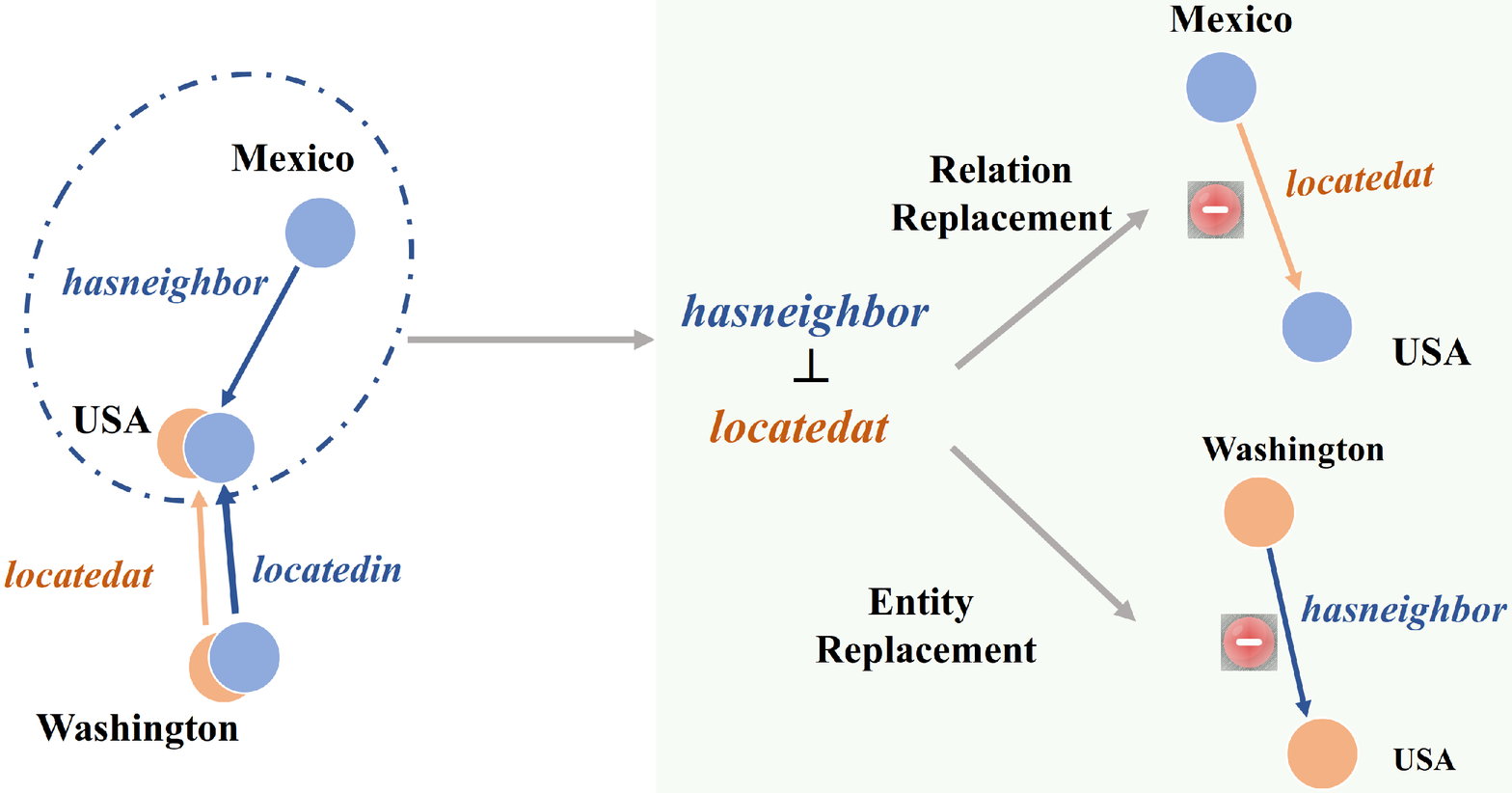}
\centering
\vspace{-0.1in}
\caption{A simple example of cross-KG negative sampling. The nodes and edges with orange color represent the target KG, and the blue colored ones denote the external KG. $\perp$ denotes the negative relationship between two relations.}
\label{fig:cross-kg}
\vspace{-0.2in}
\end{figure}

\begin{definition}[\textbf{Relation Replacement}] 
Let $S_2$ represent the set of triplets in the external KG $G_2$. For a triplet $(e_s^2, r^2, e_o^2) \in S_2$,  if we replace $r^2$ with any relation $r^1 \in N(r^2)$, the new triplet $(e_s^2, r^1, e_o^2)$ can be seen as a cross-KG negative sample. This new negative sample is made up of entities $e_s^2, e_o^2 \in \mathcal{E}^2$ and $r^1 \in \mathcal{R}^1$. $S^\prime_r$ is represented as the set of cross-KG negative samples generated by relation replacement.
\end{definition}
\vspace{-0.05in}

The intuition of relation replacement is that if a triplet $(e_s^2, r^2, e_o^2)$ is correct, but
$r^1$ and $r^2$ do not have any overlapping entity pair, i.e., no triplets can satisfy the relation $r^1, r^2$ at the same time, then the new triplet $(e_s^2, r^1, e_o^2)$ tends to be incorrect. 
\vspace{-0.05in}
\begin{example}
As shown in Fig.~\ref{fig:cross-kg} and Example~\ref{example:negative_set}, since $hasneighbor \perp locatedat$, we can replace the relation ``hasneighbor'' between the entities ``Mexico'' and ``USA'' by the negative relation ``locatedat'' to obtain a negative sample (Mexico, locatedat, USA).
\end{example}
\vspace{-0.05in}
\begin{definition}[\textbf{Entity Replacement}]
Given a triplet $(e_s^2, r^2, e_o^2) \in S_2$ and $r^1 \in N(r^2)$, if we replace $(e_s^2, e_o^2)$ with any entity pair $(e_s^1, e_o^1)$ of triplets  which satisfy $r^1$, the new triplet $(e_s^1, r^2, e_o^1)$ can be seen as a cross-KG negative sample.
\end{definition}

\begin{example}
Since (Washington, USA) satisfies the relation ``locatedat'' shown in Fig.~\ref{fig:cross-kg}, and $hasneighbor \perp locatedat$, instead of replacing the negative relation ``locatedat'', we can replace the entity pairs on the relation ``hasneighbor''. Thus, we have a new negative sample (Washington, hasneighbor, USA).
\end{example}
\vspace{-0.05in}
There may be many pairs of entities that satisfy the relation $r^1$. We only choose one pair of entities as the replacement by random sampling. The proposed cross-KG negative sampling efficiently transfers validation information from the external KG for the target KG validation. 
It is worth noting that although many negative samples generated may not be semantically close, they are still very helpful for the embedding learning. The reason is that the proposed model needs to learn from easy cases (e.g., the negative relations ``hasneighbor'' and ``hasPresident'') to difficult cases (e.g., ``hasneighbor'' and ``locatedat''). Thus, a negative sample set including many easy cases is beneficial for model learning in the early stage.  Difficult negative samples are more informative for models at the late stage. 

\vspace{-0.05in}
\subsection{Validation Optimization Framework}
\label{subsec:vof}
The goal of the proposed validation framework is to estimate the probabilities of triplets being true for \emph{the target KG}. To achieve this goal, we incorporate the confidence estimation into the proposed framework to guide the model learning by assigning larger weights to correct samples and lowering the weights of incorrect samples.

\subsubsection{Confidence Estimation in the Target KG} 
To estimate the confidence of observed triplets $S^1$ in the target KG, we propose to apply a non-linear function $\sigma(\cdot)$ (e.g., \textit{Sigmoid} function) to transform the score values from knowledge graph embedding to $[0, 1]$. This operation can be applied to any knowledge graph embedding with a \textbf{multiplicative based score function}.
Mathematically, the confidence of a triplet $ s_1 =(e^1_s, r^1, e^1_o)$ in the target KG can be measured by the output probability $P(s_1) =  \sigma(\phi (s_1))$, where $\phi$ is the score function of a knowledge graph embedding model. The initial confidence scores can be set as 1 for all the triplets.

The probabilities of corrupted samples being incorrect should be greater than or equal to the probabilities of corresponding observed samples being correct. For conventional corrupted negative samples $s^\prime_1 \in S^\prime_1$, where $S^\prime_1$ represents the set of negative samples of the target $G_1$, we can analyze their probabilities based on two cases: If the observed sample $s_1$ is correct, then the corrupted negative sample $s^\prime_1$  should be incorrect following the locally closed world assumption;  If the observed sample $s_1$ is incorrect, then the corrupted negative sample $s^\prime_1$ is still possibly incorrect. 

Based on the aforementioned two cases, the probability of negative sample $s^\prime_1$ being incorrect is greater than or equal to the probability of corresponding observed sample $s_1$ being correct. Thus, we can use the lower bound confidence, i.e., the confidence of corresponding observed triplet $P(s_1)$, as the confidence of corrupted sample $s^\prime_1$ to decrease the influence of false positive samples.


To reduce the effect caused by the spread of misinformation during training, we introduce the confidence estimation component, denoted as $\pi(\cdot)$. The samples with low confidence usually indicate the noisy ones. Thus, $\pi(\cdot)$ takes $P(s_1)$ as the input and outputs $P(s_1)$ if $P(s_1)$ is greater than a threshold $\theta$; otherwise 0. The confidence estimation component $\pi(\cdot)$ can be written as: 
\[
  \pi(P(s_1)) = 
  \begin{cases}
   P(s_1),       & \quad \text{if } P(s_1)\geq \theta\\
    0,   & \quad \text{otherwise }
  \end{cases}.
\]
We add the confidence estimation component $\pi(\cdot)$ into the loss function that derives the embedding for KG $G_1$. The loss function is defined as: 
\begin{equation}
\small
\label{eq:ob1}
\begin{aligned}
 L_{G_1} =   - \sum_{s_1 \in S_1} \sum_{s_1^\prime \in S_1^\prime}   \pi(P(s_1)) \left[ \log(P(s_1)) + \log(1- P(s_1^\prime)\right].
\end{aligned}
 \end{equation}
 
The goal would be to obtain triplet embeddings such that the confidence of positive samples in $S_1$ is high while the confidence of negative samples in $S'_1$ is low. We incorporate the confidence estimation component $\pi(P(s_1))$ into the loss so that we rely more on the samples with high confidence to derive the embeddings. 
Although the above loss function can incorporate the confidence values to decrease the influence of the misinformation, the errors in the target KG are still challenging to spot due to the limited information. To tackle this challenge, we need to incorporate an external KG to detect errors from the target KG. 

 \subsubsection{Cross-KG Negative Sampling}
To transfer information from the external KG to the target KG, we use the proposed cross-KG negative sampling method to generate cross-KG negative samples. To avoid incorporating false negative samples, the negative sample generation is only conducted based on the triplets from the external KG. In other words, given a triplet from the external KG, we use entity replacement and relation replacement strategies to generate cross-KG negative samples, which guarantee the information transfer between two KGs. 
 
The generated cross-KG negative samples can be added to the representation learning framework following the similar strategy adopted by conventional negative sampling methods. The loss function that infers the embeddings of  the external KG $G_2$ is defined as follows:
  \begin{equation}
  \small
  \label{eq:ob2}
 \begin{aligned}
      L_{G_2} = & - \sum_{s_2 \in S_2}\Big[ \log(P(s_2)) + \sum_{s^\prime_2 \in S^\prime_2}  \log(1-P(s^\prime_2)) \\ 
      & + \sum_{s^\prime_e \in S^\prime_e} \log(1-P(s^\prime_e)) + \sum_{s^\prime_r \in S^\prime_r}\log(1-P(s^\prime_r)) \Big].
      \end{aligned}
\vspace{-0.1in}
\end{equation}
For each triplet $s_2$, we generate its corresponding negative sample set $S'_e$ , $S'_r$ and $S'_2$. In Eq.~\ref{eq:ob2}, $S'_2$ denotes conventional negative samples, $S'_e$ denotes the negative samples generated from entity replacement, and $S'_r$ denotes the negative samples generated from relation replacement.  Note that the confidence estimation component is only used on the \emph{target} graph which has errors to be detected. 

\subsubsection{Objective Function}
Our ultimate goal is to validate the target KG by jointly learning over both target and external KGs. Thus, the final objective function is a combination of two parts:  
\begin{equation}
\label{eq:final}
L_{final} = L_{G_1} + \lambda L_{G_2}.
\end{equation}
Here, $\lambda$ controls the balance between $L_{G_1}$ and $L_{G_2}$, and we simply set the value of $\lambda$ as 1. In practice, the proposed framework is not sensitive to this hyper-parameter, and we investigate the impact of $\lambda$ in Subsection~\ref{experiment:second_loss}.

\vspace{-0.1in}
\section{Experiments}
\label{sec:exp}


In this section, we first introduce the datasets used in the experiments, and then present the settings of the experiments, including baselines, evaluation measures and implementation details. Finally, we show the effectiveness of the proposed framework and qualitatively analyze the insights obtained from the experiments. 

\subsection{Knowledge Graphs \& Datasets}\label{subsection:kg_d}

In this subsection, we first introduce the knowledge graphs in the medical and general domains. In each domain, there are two KGs: one of the KGs is the target KG, which is used for validation purpose and the other is the external KG which is used to provide auxiliary information. 


\vspace{-0.05in}
\subsubsection{Medical Domain}
\label{sec:medical}
\textbf{1) Target KG.} KnowLife~\cite{ernst2015knowlife} is a huge medical knowledge repository. It is automatically constructed based on scientific literature, and thus may contain errors. Hence the KG KnowLife is used as the target KG for validation. We crawled from their public website\footnote{http://knowlife.mpi-inf.mpg.de/} and obtained a KG, including 656,607 entities, 25 relations and 2,005,369 triplets.  \textbf{2) External KG.} SemMedDB~\cite{kilicoglu2011constructing} is a large-scale human-curated biomedical knowledge graph whose facts are more likely to be correct. Therefore, SemMedDB is used as the external knowledge graph to provide auxiliary validation information. The SemMedDB contains 154,216 entities, 63 relations and 1,804,054 triplets.  Based on the target KG KnowLife and the external KG SemMedDB, we construct a dataset with synthetic errors for the validation of the proposed framework. 

\smallskip
$\bullet$ \textbf{Dataset}.  Given a positive triplet $(e_s, r, e_o)$ from KnowLife, we randomly corrupt either head or tail entity to form a negative triplet $(e_s^\prime, r, e_o)$ or $(e_s, r, e_o^\prime)$. More specifically, we randomly sample 1,250 observed triplets and denote the collection of these triplets as $\mathcal{L}$. Negative samples are generated correspondingly, and the set is denoted as $\mathcal{G}^-$. $\mathcal{G}^- \nsubseteq \text{KnowLife}$, and its size is also 1,250. 
In this dataset, we treat observed triplets  as positive and generated triplets $\mathcal{G}^-$ as negative, and the positive/negative labels are the groundtruth. When we use the proposed framework, 20\% of triplets randomly selected from $\mathcal{L}$ and $\mathcal{G}^-$ are used to select the best parameters, and the remaining 80\% of data are used for evaluation. 

\subsubsection{General Domain}
 \textbf{1) Target KG.} NELL~\cite{mitchell2018never} is a huge and growing knowledge graph, which is constructed by automatic information extraction approaches on unstructured content. Since the information extractors are imperfect, it is inevitable to extract noisy and incorrect information. Thus, we use NELL as the target KG for validation. We first select the entities that appear in at least 5 triplets. Following the similar setting of knowledge graph completion, the reverse relations of NELL are then removed. Moreover, we remove the triplets of relation ``generalizations'' that are ambiguous to verify and are dominating compared with the other relations. After these processing steps, we obtain the target KG, denoted as NELL-314, which contains 13,965 entities, 314 relations, and 148,001 triplets. \textbf{2) External KG.} YAGO~\cite{yago} is a large-scale knowledge graph constructed on semi-structured content, and its accuracy is estimated as 95\%\footnote{https://en.wikipedia.org/wiki/YAGO\_(database)}. Here, YAGO is used as the external knowledge graph to assist the validation on the target KG NELL-314. To preprocess YAGO, we first remove the relations without any overlapping entities with NELL-314, and then select the entities that appear in at least 5 triplets. This results in the external knowledge graph YAGO$_e$ with 242,281 entities, 27 relations, and 2,324,069 triplets.



$\bullet$ \textbf{Dataset}. \textbf{1) Human-annotated Labels.} To fairly evaluate the performance of the proposed framework, we manually label triplets that are randomly selected from the target KG NELL.
We first employ two participants to manually label each triplet (e.g., true/positive or false/negative) by cross-checking with the information presented on Wikipedia. 
If the labels of a triplet provided by the two participants are different, then this triplet will be assigned to the third one for further investigation. Majority voting is applied to decide the final label of each triplet. 
Following this procedure, we label 2,500 triplets, which is divided into two sets, i.e., the set of positive samples $\mathcal{L}^+$ and the set of negative samples  $\mathcal{L}^-$.  For both, the size is 1,250. Note that $\mathcal{L}^+, \mathcal{L}^-  \subset \text{NELL}$. \textbf{2) Synthetic Dataset}. 
We first construct a dataset with synthetic errors based on NELL, referred to as synthetic dataset. 
We follow the same procedure used to generate synthetic errors for our datasets in medical domain as introduced in Section~\ref{sec:medical}.  In this way, we generate a set of negative samples with size 1,250, denoted as $\mathcal{G}^-$. Then 20\% of the triplets randomly selected from $\mathcal{L}^+$ and $\mathcal{G}^-$ are used to select the best parameters, and the remaining 80\% of data are used for evaluation. 
 \textbf{3) Real Dataset}\footnote{Since there is no publicly available human labeled  datasets for the knowledge graph validation task, we will release this dataset to encourage the future research on this important problem.} In addition to a dataset with synthetic errors, we also construct a dataset whose errors are real and annotated by human. In this real dataset, we do not synthesize any erroneous fact for NELL. We use $\mathcal{L}^+$ and $\mathcal{L}^-$ as the groundtruth positive and negative cases for evaluation. Similarly, 20\% of data randomly chosen from them are used for the selection of parameters, and 80\% for performance evaluation.


\vspace{-0.05in}
\subsection{Experiment Settings}
\emph{Baseline Approaches.}
We compare the proposed framework \textsf{CrossVal} with the state-of-the-art basic embedding models, including DistMult~\cite{yang2014embedding}, TransE~\cite{bordes2013translating},  ComplEx~\cite{trouillon2017knowledge}, Analogy~\cite{liu2017analogical} and SimplE~\cite{kazemi2018simple}.  Moreover, we add CKRL~\cite{xie2018does}, which employs the score function of TransE and adds the  confidence estimation of triplets learned from the internal structural information to its objective function, as a baseline.  We also add a multi-KG embedding method PTransE~\cite{zhu2017iterative} as a baseline, whose score function is same as that of TransE. Note that multi-KG embedding method PTransE~\cite{zhu2017iterative}   assumes that different KGs use the same ontology, and thus the method is not directly applicable to the setting of this paper. We remove the matching loss of PTransE on predicates and only keep matching loss on entities. All of these baseline methods use the target KG and the external KG for training purpose, and we further investigate their  performance on the target KG only in Subsection~\ref{subsec: ab}.  In this paper, we do not compare our proposed models with rule-based models because many triplets in the given test sets are not covered by the rules. 


\emph{The Proposed Approaches.} The proposed framework \textsf{CrossVal} can be easily applied to the state-of-the-art KG embedding methods that are based on multiplicative score functions. Thus, we propose four enhanced models: \textsf{CrossVal}$_{Dist}$,  \textsf{CrossVal}$_{Comp}$, \textsf{CrossVal}$_{Ana}$, and \textsf{CrossVal}$_{Simp}$  based on DistMult, ComplEx, Analogy and SimpIE embedding approaches, respectively. TransE, CKRL and PTransE are translational based models, and thus they are not applicable to the proposed framework. 

\smallskip
$\bullet$ \textbf{Evaluation Measures.}
We use ranking measures to evaluate the performance of all the compared approaches. The ranks can be obtained according to the scores outputted by the models for the triplets in the KG. A smaller score for a triplet means the lower possibility of it being true. We rank all the triplets in the target KG according to their scores in ascending order. The top ranked triplets are more likely to be wrong. To fairly evaluate the performance of the KG validation, we use three evaluation measures: \textbf{Mean Rank}, \textbf{Precision@K}, \textbf{Recall}. To mathematically introduce the measures, we define the evaluation set as $D$, which consists of the positive triplet set $D^+$ and negative triplet set $D^-$. For the $i$-th triplet, the $rank_i$ represents its rank in the evaluation set $D$.


\emph{Mean Rank.} We follow previous work~\cite{melo2017detection} to evaluate the KG validation. We use the \textit{Mean Raw Rank}, as well as its filtered variation, named \textit{Mean Filter Rank}, which filters out correctly higher ranked predictions. 
The measures are defined as follows:
\begin{equation}
\small
\text{Mean Raw Rank} = \frac{1}{|D^-|} \sum_{i=1}^{|D^-|} rank_i, 
\normalsize
\end{equation}
\vspace{-0.1in}
\begin{equation}
\small
\text{Mean Filtered Rank}  = \frac{1}{|D^-|} \sum_{i=1}^{|D^-|} rank_i - i.
\normalsize
\end{equation}
When we calculate the \textit{Mean Filter Rank}, we first order the incorrect triplets by their ranks in ascending order. For instance, if the ranks of incorrect triplet set $D^-$ are $\{1,2,4,7\}$, then the filtered sequence of ranks is $\{0,0,1,3\}$, and its \textit{ Mean Filter Rank} is 1.
The smaller both \textit{Mean Raw Rank} and \textit{Mean Filter Rank} are, the better the performance is. 

\emph{Recall of Ranking.} Considering that the number of verified incorrect triplets is known, the rank of the incorrect triplets can be thought as correct if its rank is in the range $[1,|D^-|]$. Towards this end, we propose a metric called \textit{Recall of Rank}, denoted as \textit{Recall}, which is equivalent to  the Precision@$|D^-|$. 
\begin{equation}
\small
\begin{aligned}
\text{Recall} = \frac{1}{|D^-|} \sum_{{i=1}}^{|D^-|} \mathbbm{1}_{[1,|D^-|]}~(rank_i). 
\end{aligned}
\normalsize
\end{equation}


\smallskip
$\bullet$ \textbf{Implementation Details.}
We implement all the deep learning based baselines and the proposed framework with PyTorch 1.2. We use Nvidia Titan Xp GPU server and train our proposed framework using Adam~\cite{kingma2014adam} for optimization. We select the hyper-parameters following the same way used by  \cite{zhao2019scef, transH, kazemi2018simple, liu2017analogical}. The overall learning rate for the proposed framework is selected among $\{0.0001, 0.0005, 0.001, 0.01\}$, which is fixed during the learning procedure. The embedding dimensionality is selected among $\{32, 64, 128, 256\}$, and the mini-batch size is selected among $\{64, 128, 256, 512, 1024\}$. We use the regularization term (i.e., l2 norm with the coefficient 0.001). The threshold $\theta$ is set to 0.5 and  $\lambda$ is set to 1. The optimal configurations of the proposed approaches are optimized on the 20\% labeled evaluation set as described in Subsection~\ref{subsection:kg_d}. 

\subsection{Performance Comparison}
\label{subsec:peformance}

Table~\ref{tab: medical} and \ref{tab: nell-341} show the performance of different approaches across different domains. We can observe that the proposed framework \textsf{CrossVal} achieves the best results in terms of all the evaluation metrics. 

\subsubsection{Medical Domain} 
As shown in Table~\ref{tab: medical}, on the medical datasets, Analogy achieves the best \emph{Recall} value among all the multiplicative embedding baselines. It is because Analogy explicitly models analogical structures in multi-relational embedding and is able to capture more information of KG. Compared with multiplicative embedding baselines, translational methods achieve relative lower recall values and precision values. The reason is that it is difficult for TransE (or CKRL, PTransE) to capture 1-to-N relationships~\cite{bordes2013translating} which widely exist in the medical KGs. To bridge two KGs, PTransE incorporates the alignment loss and achieves more than 3\% improvement on \textit{Recall} compared with TransE. Due to unavoidable extraction errors existing in the target KG, CKRL which is equipped with the confidence estimation procedure achieves a higher recall value compared with TransE and PTransE. Compared with these baselines,  the proposed framework achieves improvement in terms of all the metrics. For example, CrossVal$_{Ana}$ achieves a recall value as high as 0.890, which is about 12.66\% improvement compared with that of Analogy.

\begin{table}[htb]
\vspace{-0.1in}
\centering

\caption{The performance comparison of different methods on the Medical KG datasets.} 
\label{tab: medical}
\vspace{-0.1in}
\resizebox{0.9\linewidth}{!}{
\begin{tabular}{c|c|c|c|c|c|c|c}
\toprule

\multirow{2}{*}{Category}&\multirow{2}{*}{Method} & \multirow{2}{*}{Recall}&
\multicolumn{2}{c|}{Mean Rank}
&
\multicolumn{3}{c}{Precision@K} \\

\cline{4-8}
&&&Filter&Raw & 
100  & 200 & 500\\
\midrule
\midrule
\multirow{4}{*}{Multiplicative}&\textsf{DistMult} & 0.765 & 140 & 638 & \textbf{1.000} & 0.990 & 0.960 \\

&\textsf{ComplEx} & 0.774 & 142 & 641 & \textbf{1.000} & 0.995 & 0.92  \\
&\textsf{Analogy} & 0.790 & 122 & 620 & \textbf{1.000} & 0.995 & 0.978  \\
&\textsf{SimplE} & 0.752 & 156 & 654 & \textbf{1.000} & 1.000 & 0.960   \\
\cline{1-8}
\multirow{3}{*}{Translational}&\textsf{TransE} & 0.704 & 237 & 735 & 0.990 & 0.940 & 0.838 \\
&\textsf{CKRL} & 0.751 & 184 & 682 & 0.990 & 0.955 & 0.880 \\
&\textsf{PTransE} & 0.729 & 198 & 696 & 0.980 & 0.965 & 0.874  \\

\cline{1-8}

\multirow{4}{*}{Proposed}&\textsf{CrossVal}$_{Dist}$ & 0.859 & 69 & 567 & \textbf{1.000} & \textbf{1.000} & 0.996   \\
&\textsf{CrossVal}$_{Comp}$ &  0.884 & 54 & 552 & \textbf{1.000} & \textbf{1.000} & \textbf{1.000}  \\
&\textsf{CrossVal}$_{Ana}$ & \textbf{0.890} & \textbf{53} & \textbf{551} & \textbf{1.000} & \textbf{1.000} & 0.996   \\
&\textsf{CrossVal}$_{Simp}$ & 0.863 & 68 & 567 & \textbf{1.000} & \textbf{1.000} & 0.994  \\
\bottomrule
\end{tabular}
}

\vspace{-0.1in}
\end{table}

\subsubsection{General Domain} 
\begin{table*}[htb]
\centering
\caption{The performance comparison of different methods in the Synthetic and Real settings on the NELL-314. }
\vspace{-0.1in}
\label{tab: nell-341}

\resizebox{0.75\textwidth}{!}{
\begin{tabular}{c|c|c|c|c|c|c|ccc|c|c|c|c|c}
\toprule
&&\multicolumn{6}{c}{Synthetic} && \multicolumn{6}{c}{Real}\\
\cline{3-8}
\cline{10-15}
\multirow{2}{*}{Category}&\multirow{2}{*}{Method} & \multirow{2}{*}{Recall}&
\multicolumn{2}{c|}{Mean Rank}
&
\multicolumn{3}{c}{Precision@K}  && \multirow{2}{*}{Recall}&
\multicolumn{2}{c|}{Mean Rank}
&
\multicolumn{3}{c}{Precision@K}\\

\cline{4-8}
\cline{11-15}
&&&Filter&Raw & 
100  & 200 & 500 &&&Filter&Raw & 100  & 200 &  500\\
\midrule
\midrule
\multirow{4}{*}{Multiplicative}&\textsf{DistMult} & 0.842 & 86 & 585 & 0.950 & 0.900 & 0.898 && 0.584 & 395 & 894 & 0.660 & 0.670 & 0.604 \\

&\textsf{ComplEx} & 0.871 & 78 & 577 & 0.950 & 0.895 & 0.878 && 0.570 & 403 & 902 & 0.690 & 0.680 & 0.598  \\
&\textsf{Analogy} & 0.862 & 78 & 576 & 0.940 & 0.905 & 0.902 && 0.587 & 376 & 874 & 0.680 & 0.650 & 0.600 \\
&\textsf{SimplE} & 0.879 & 48 & 547 & \textbf{1.000} & 0.995 & 0.984 && 0.591 & 380 & 879 & 0.680 & 0.700 & 0.596  \\
\cline{1-15}

\multirow{3}{*}{Translational}&\textsf{TransE} & 0.884 & 44 & 542 & 1.000 & 1.000 & 0.986 && 0.548 & 448 & 946 & 0.540 & 0.555 & 0.552 \\
&\textsf{CKRL} & 0.889 & 40 & 539 & \textbf{1.000} & \textbf{1.000} & 0.980 && 0.531 & 442 & 941 & 0.690 & 0.630 & 0.586  \\
&\textsf{PTransE}& 0.854 & 74 & 572 & \textbf{1.000} & \textbf{1.000} & 0.986 && 0.609 & 371 & 870 & 0.590 & 0.625 & 0.580\\

\cline{1-15}
\multirow{4}{*}{Proposed}&\textsf{CrossVal}$_{Dist}$ & \textbf{0.923} & 30 & 531 & \textbf{1.000} & \textbf{1.000} & 0.982  && 0.641 & 305 & 803 & 0.780 & 0.720 & \textbf{0.704}  \\
&\textsf{CrossVal}$_{Comp}$ &  0.921 & \textbf{28} & \textbf{528} & \textbf{1.000} & 0.995 & \textbf{0.988}  && 0.639 & 306 & 805 & 0.740 & 0.695 & 0.700  \\
&\textsf{CrossVal}$_{Ana}$ & \textbf{0.923} & 29 & 530 & 0.990 & 0.990 & 0.986  && \textbf{0.645} & 299 & 798 & \textbf{0.800} & \textbf{0.765} & \textbf{0.704}  \\
&\textsf{CrossVal}$_{Simp}$ & 0.920 & 31 & 531 & \textbf{1.000} & 0.995 & 0.980  && 0.634 & \textbf{298} & \textbf{797} & \textbf{0.800} & 0.740 & 0.692  \\
\bottomrule
\end{tabular}
}

\vspace{-0.1in}
\end{table*}

\textbf{Synthetic Dataset}. 
  Compared with other KG embedding models, the \emph{Recall} value of DistMult is the lowest, as shown in Table~\ref{tab: nell-341}. It is because DisMult is based on simple multiplicative score function which cannot model the anti-symmetric relations.
    Although CompIEx, Analogy and SimpIE are also multiplicative models, their performance is much better than that of DisMult. The reason is that they modify the score functions to model more complex relationships. Especially, the \textit{Recall} of SimpIE improves 4.4\% compared with that of DistMult.
    Different from multiplicative based models, TransE and CKRL are translational methods, which use the additive score functions and employ the margin loss function to increase the margin between positive and negative samples. 
Although the score function of TransE is very simple, its margin loss function is robust to noise, which leads to high \textit{Recall} of the model (i.e., 88.4\%). 
    PTransE and CKRL are built upon TransE.  CKRL considers the confidence value estimation of triplets, which further improves the performance compared with TransE.
    From all the baselines, we can observe that translational methods outperform multiplicative based models on the synthetic error detection. Though the proposed models \textsf{CrossVal}$_{Dist}$, \textsf{CrossVal}$_{Comp}$, \textsf{CrossVal}$_{Ana}$ and \textsf{CrossVal}$_{Simp}$ are multiplicative models, their performance is better than not only their basic models but also translational methods. The improvement is more than 4\% on \textit{Recall} and more than 30\% on \textit{Mean Filter Rank}. 
  \textbf{Real Dataset.}  As observed from Table~\ref{tab: nell-341}, the \textit{Recall} of all the methods significantly decreases around 30\% compared with that on the synthetic dataset. The performance gap between two datasets demonstrates that the real errors are more difficult to be detected than synthetic ones.
    The \textit{Recall} and \textit{Mean Rank} values of CKRL are the best among all the baselines on the synthetic dataset. However, on the real dataset, it  achieves lowest \textit{Recall} values compared with other baselines. The reason is that CKRL estimates the confidence of triplets by the margin between observed triplets and generated errors. However, the real errors are difficult to tell and hence the scores of real errors are usually higher than those of the generated negative samples accordingly. Based on the mechanism of CKRL,  the confidence of real errors will be increased and this further hurts the performance. Different from CKRL, the proposed framework only relies on the observed samples when conducting the confidence estimation. Thus, such mechanism leads to more accurate confidence estimation compared with CKRL's. Besides, the proposed framework \textsf{CrossVal} uses the cross-KG negative sampling to transfer information from YAGO$_e$ to NELL-314. Our framework \textsf{CrossVal} is general and its score function component can be replaced by any multiplicative score function. Thus, the framework can benefit from the improvements in the score functions and hence further improve its performance.  As observed in Table~\ref{tab: nell-341}, the proposed models achieve significant improvement--more than 7\% on \textit{Recall} compared with their corresponding approaches on the real dataset.

\vspace{-0.1in}
\subsubsection{Comparison between Synthetic \& Real Data} 
 On the synthetic dataset, all the injected incorrect triplets are generated by corrupting head or tail of observed triplets. It means that  corresponding correct samples are contained in the dataset. Thus, the correct information can guide the models to find incorrect triplets. As observed from Table~\ref{tab: nell-341}, all the approaches can achieve relatively good performance on the synthetic dataset.
    However, on the real dataset, we do not have enough information for validating incorrect triples, which limits the models for spotting more errors. It can also be validated from Table~\ref{tab: nell-341} that the difference between synthetic and real errors further leads to a general 30\% gap in terms of \textit{Recall} between two datasets. This shows that the synthetic dataset may not include the key challenge of the KG validation task. 

\begin{figure*}[htbp]
\centering
\begin{minipage}[t]{0.3\textwidth}
\centering
\includegraphics[width=1.6in]{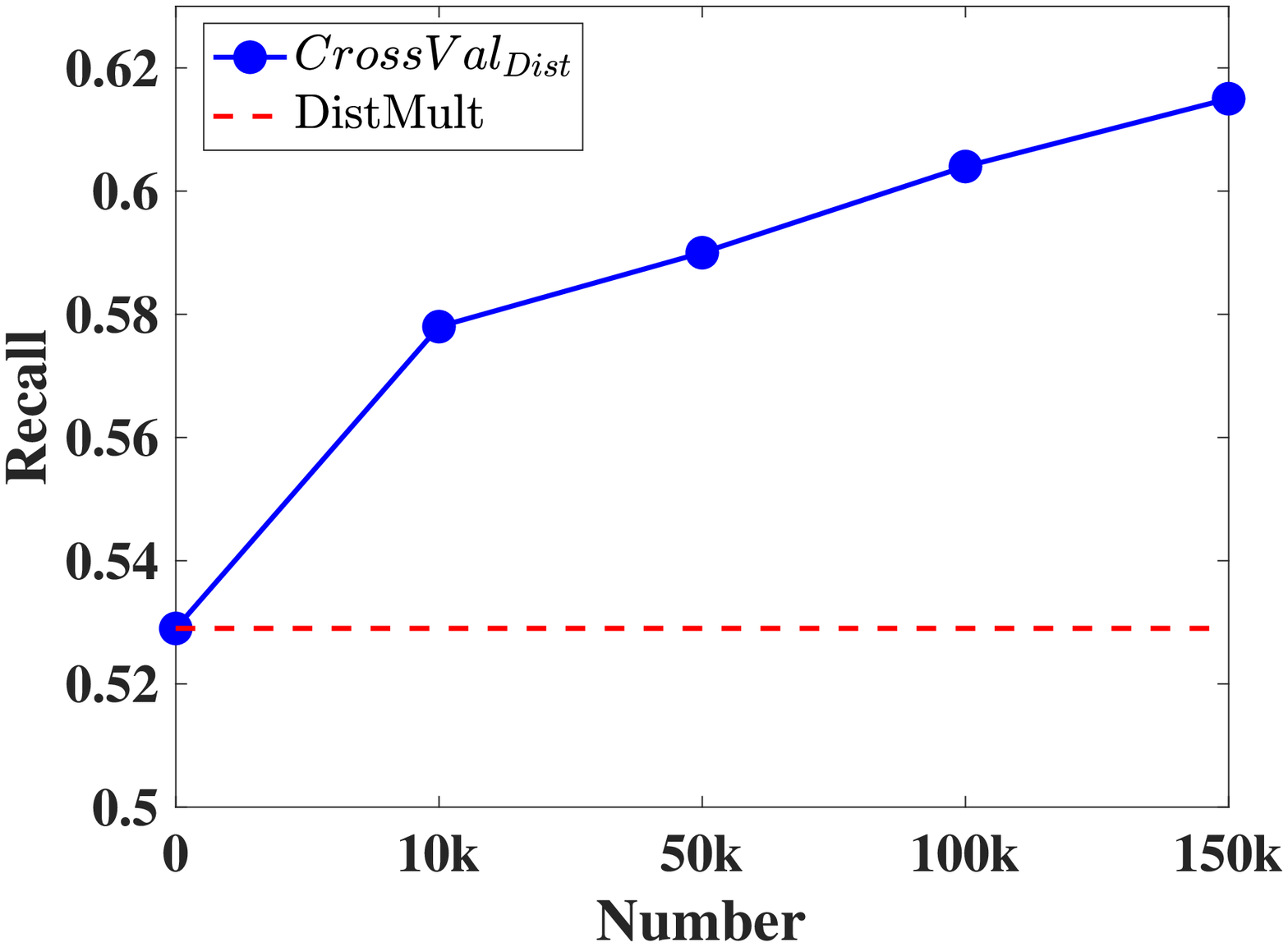}
\vspace{-0.1in}
\caption{The performance w.r.t. size of the external KG.}
\label{fig:external_data}
\end{minipage}
\hspace{0.1in}
\begin{minipage}[t]{0.3\textwidth}
\centering
\includegraphics[width=1.6in]{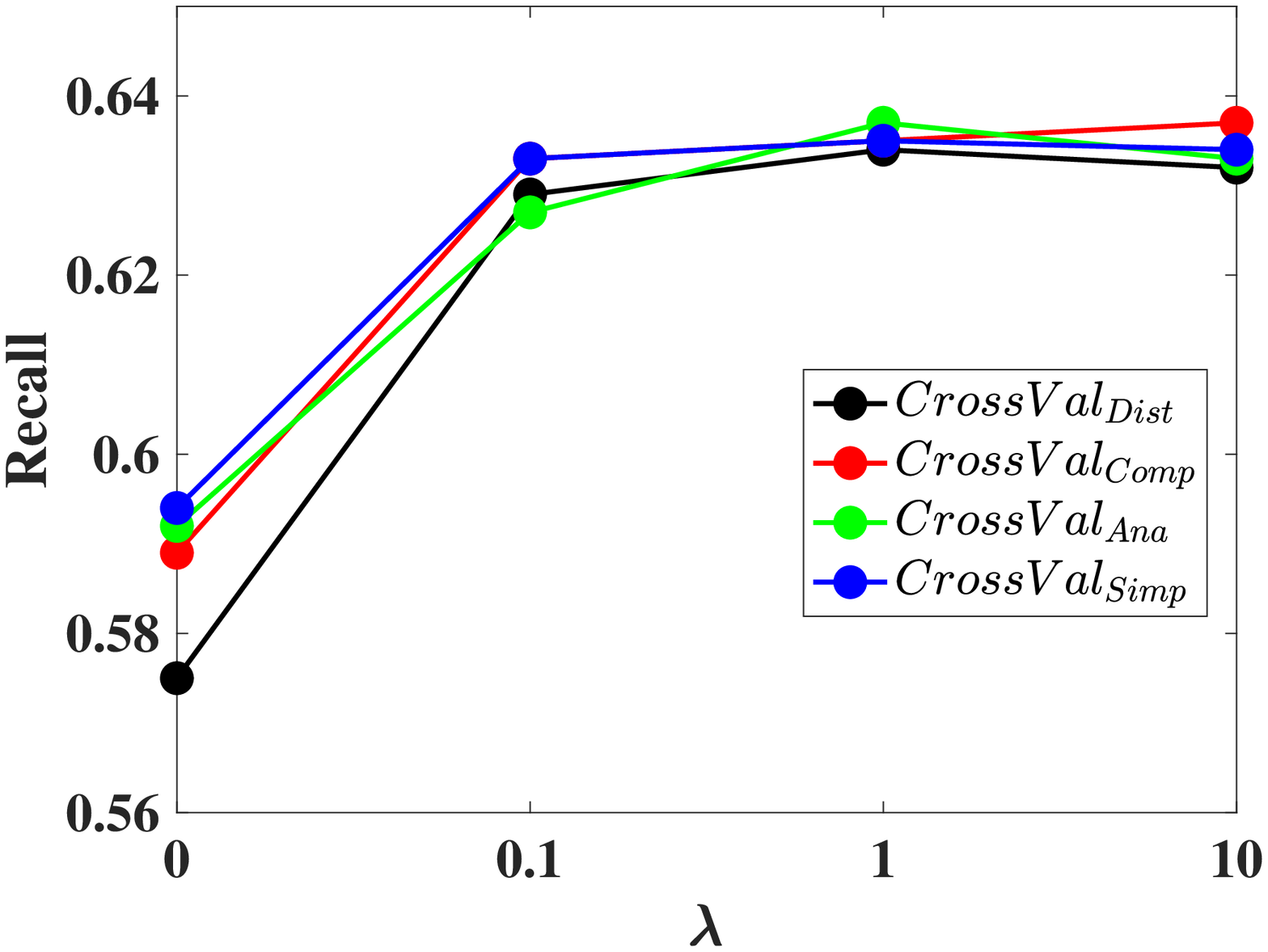}
\vspace{-0.1in}
\caption{The performance w.r.t. different $\lambda$.}
\label{fig:lambda}
\end{minipage}
\hspace{0.1in}
\begin{minipage}[t]{0.3\textwidth}
\centering
\includegraphics[width=1.6in]{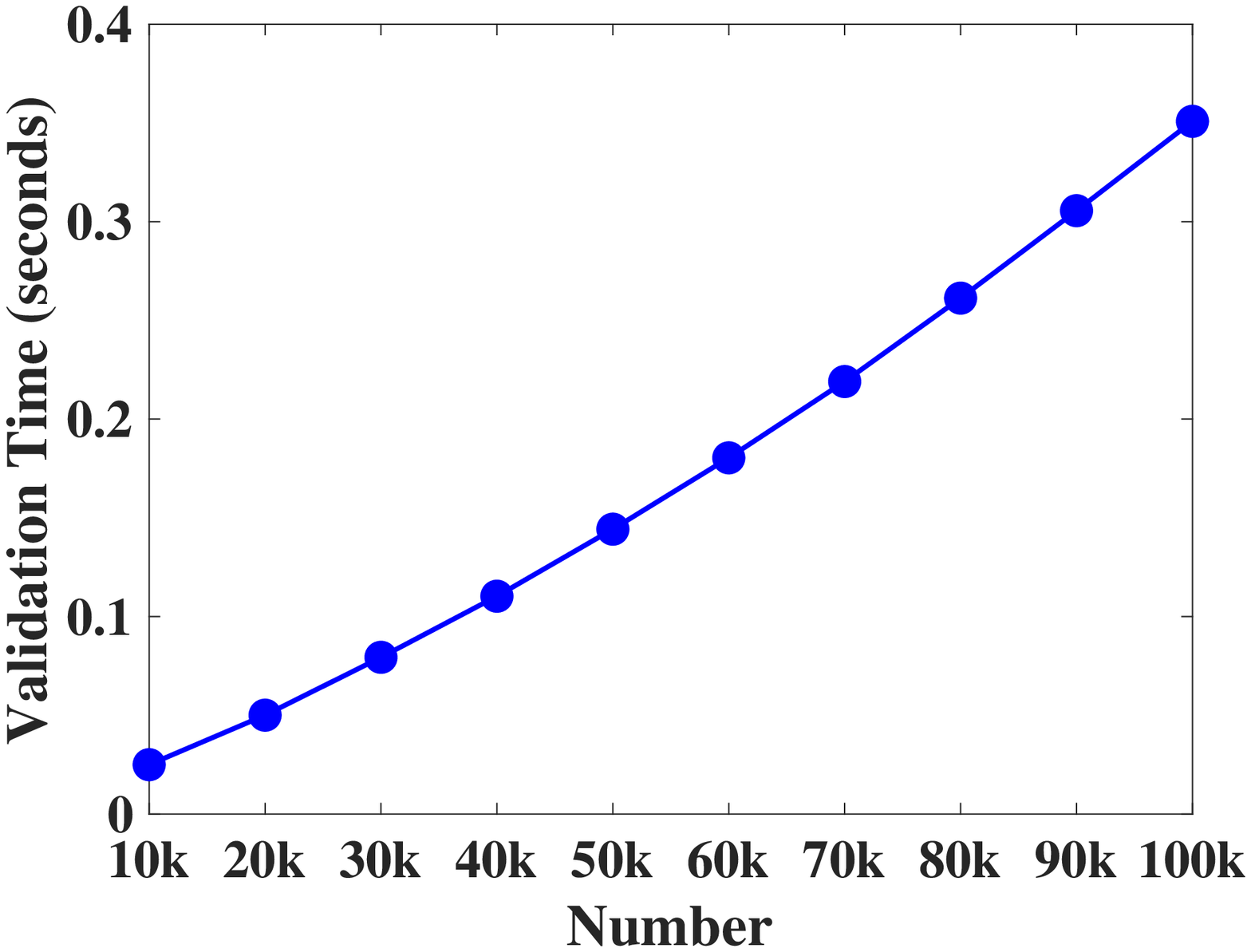}
\vspace{-0.1in}
\caption{ The validation time of CrossVal$_{Dist}$ w.r.t. \# triplets}
\label{fig:validation_time}
\end{minipage}
\vspace{-0.1in}
\end{figure*}

\vspace{-0.1in}
\subsection{Analysis of Real Errors}
      To  understand what kinds of incorrect triplets that the proposed framework is able to detect,  we first check the detected incorrect triplets on the NELL official website\footnote{http://rtw.ml.cmu.edu/rtw/}, which provides the extract patterns and source links for the extracted triplets. We then roughly classify those errors into three categories: \textit{entity error}, \textit{relation error} and \textit{propagation error}.  The entity errors usually occur during the procedure of named-entity recognition (NER). When the extracted entities are correct, the triplet still have a probability to be wrong, which stems from the incorrect relation between a pair of entities. This kind of errors is relation errors. 
    The propagation error refers to the knowledge reasoning errors originated from rules or other mechanisms. We conduct case studies using representative examples for each category of errors and provide the detailed reasons by analyzing the corresponding information sources of examples. 

$\bullet$ \textbf{Entity Errors} (\emph{Fred\_Goodwin, ceoof, Scotland}). According to the definition of relations provided by the NELL official website, ``ceoof'' specifies that a particular person is the CEO of a particular company\footnote{http://rtw.ml.cmu.edu/rtw/kbbrowser/predmeta:ceoof}.
Therefore, this triplet means \emph{Fred Goodwin} is the CEO of \emph{Scotland}. We check the information source and find that the related textual statement is \textit{``Former Royal Bank of \underline{Scotland} boss \underline{Fred Goodwin} has had his knighthood removed.''}\footnote{https://www.bbc.com/news/uk-politics-16821650} The correct triplet based on this sentence should be (\emph{Fred\_Goodwin, ceoof, Royal\_Bank\_of\_ Scotland}). The algorithm mistakenly recognizes the object entity ``Royal Bank of Scotland'' as  ``Scotland'' and this leads to a incorrect triplet. This kind of errors is very common in the current KG.

$\bullet$ \textbf{Relation Errors}
 (\emph{two\_million\_people, isshorterthan, washington}). The meaning of ``isshorterthan'' indicates that object entity has a greater height than subject entity\footnote{http://rtw.ml.cmu.edu/rtw/kbbrowser/predmeta:isshorterthan}. Thus, this triplet means that the height of Washington is greater than that of two million people. 
 This triplet is extracted  from the following news:
\textit{``\underline{WASHINGTON}, Dec 31 (Reuters) - Over \underline{two million people} have  enrolled in health insurance plans through the federally run HealthCare.gov and state healthcare enrollment websites, a U.S. administration official said on Tuesday.''}\footnote{https://www.huffpost.com/entry/obamacare-signups\_n\_4524556} 
    The NER tool can correctly recognize the entities ``washington'' and ``two\_million\_people''. However, when extracting the relation between them, the algorithm considers that ``over'' is the key word, and mistakenly assigns the relation ``isshorterthan'' between them.
    This type of errors is difficult to be detected using type constraints because the types of both ``washington'' and ``two\_million\_perople'' are the same: \emph{Person}. This triplet does not violate the constraints for the type ``isshorterthan''. 
Since one entity usually has multiple correct entity types, how to correctly predict the type of the entity based on the specific context in the information source is very important for relation extraction task.

$\bullet$ \textbf{Propagation Errors}  (\emph{two\_million\_people, isshorterthan, dc}) and (\emph{washington, istallerthan, two\_million\_people}).
We provide two representative examples to illustrate the propagation errors through entities and relations, respectively. 
    In the previous example, there is an incorrect triplet  (\emph{two\_million\_people, isshorterthan, washington}). Since ``Washington'' is usually equivalent to ``DC'', the error is propagated by the extraction algorithm, which produces a new incorrect triplet (\emph{two\_million\_people, isshorterthan, dc}).
    The second example shows the error caused by reasoning on relations. The relation ``istallerthan'' is the inverse relation of ``isshorterthan''. Thus, the triplets with the relation ``istallerthan'' can be created by exchanging the subject and object entities' orders of triplets with the relation ``isshorterthan''. In such a way, the error information can be propagated over the KG.

\subsection{Ablation Study}

\label{subsec: ab}

    \textbf{Importance of External Information.} 
    In this study, we aim to analyze the performance changes of baseline methods incurred by incorporating an external knowledge graph. Note that we use conventional negative sampling instead of the proposed  cross-KG negative sampling method to generate negative samples.

\begin{figure}[hbt]
\vspace{-0.2in}
\subfloat[Recall]{
\vspace{-0.2in}
\begin{minipage}{.21\textwidth}
\includegraphics[height=1.0in]{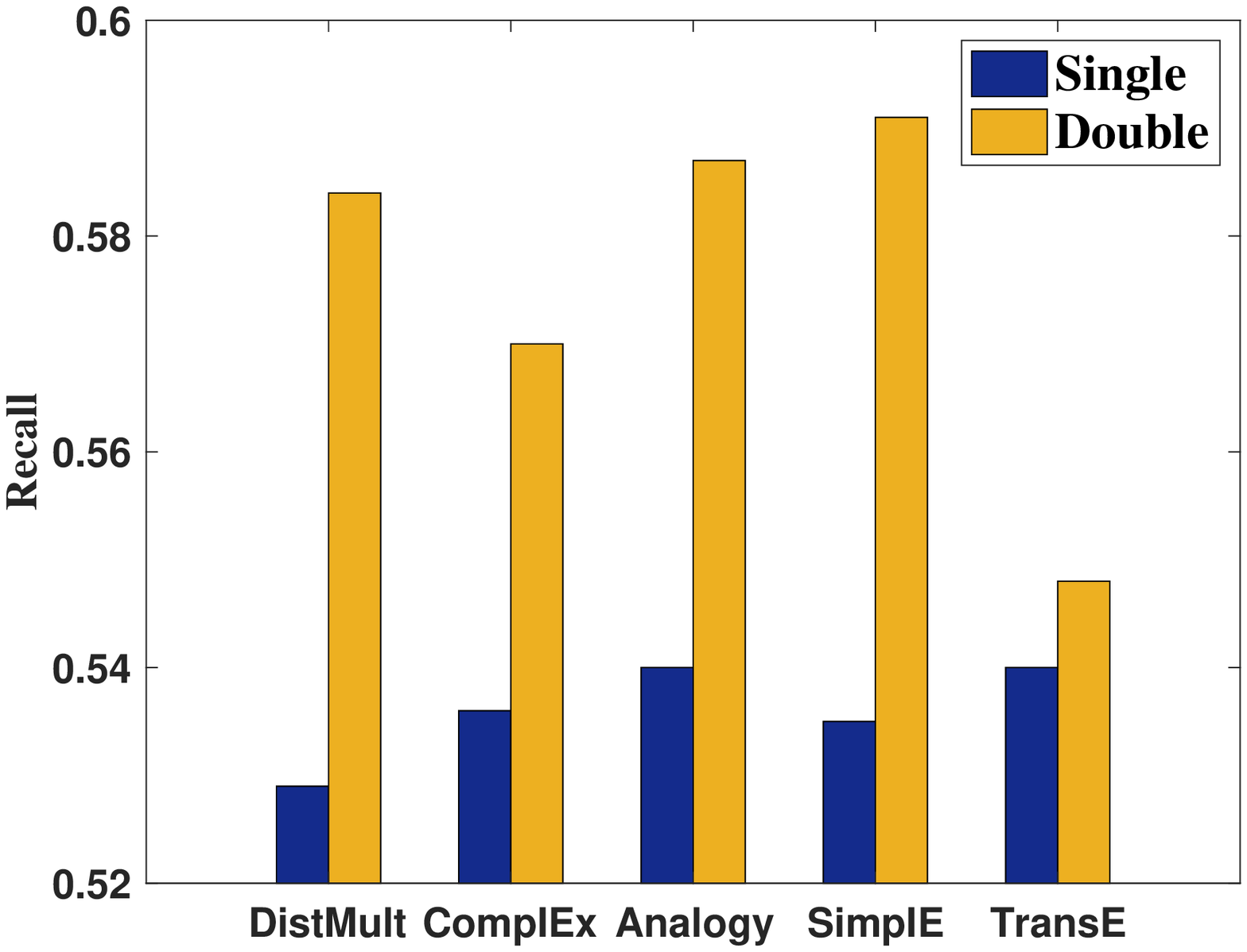}
\label{subfig:double_recall}
\end{minipage}}
\vspace{-0.1in}
\subfloat[Filter Mean Rank]{
\begin{minipage}{.22\textwidth}
\includegraphics[height=1.in]{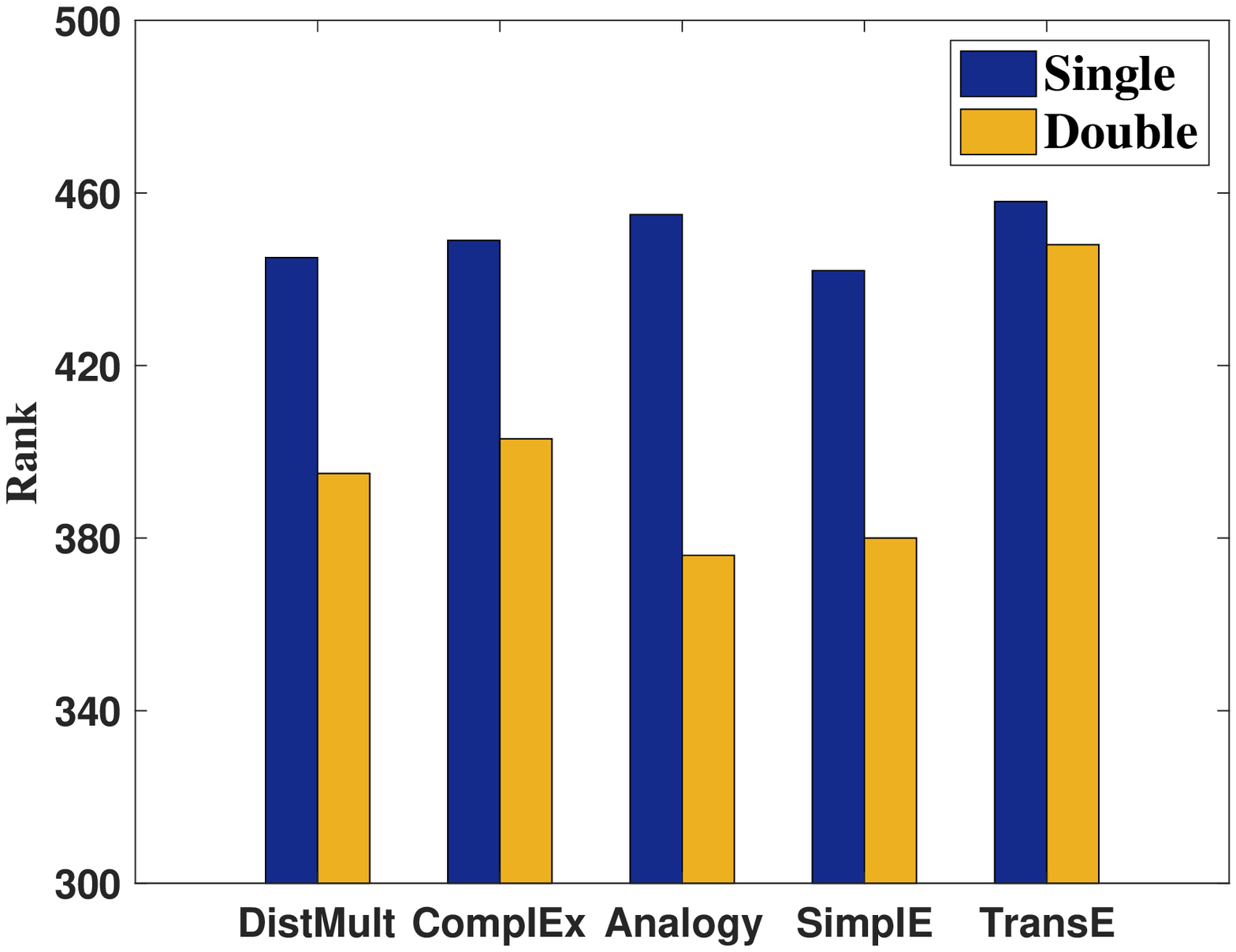}
\label{subfig:double_rank}
\end{minipage}}
 \caption{The performance comparison of methods on single and two knowledge graphs.}\label{fig:single_double}
 \vspace{-0.1in}
\end{figure}

    Fig.~\ref{fig:single_double} shows the \textit{Recall} and \textit{Filter Mean Rank} of all the baselines (except CKRL\footnote{The score function of CKRL is similar to that of TransE, and CKRL's performance is highly related to the tuning of hyper-parameters. Thus, we only show the performance of TransE as a representative in this experiment.}) when incorporating the external KG YAGO$_e$. ``Single'' denotes the KG embedding methods on the target KG NELL-314, and ``double'' represents that the embedding methods are learned on both YAGO$_e$ and NELL-314. We can observe that the \textit{Recall} and \textit{Mean Filter Rank} of every method improves when incorporating the external YAGO$_e$ . This again validates the importance of external information in the error detection.

    \textbf{Importance of Confidence Estimation Component.} 
    The confidence estimation step not only helps the KG embedding methods to lower the influence of errors, but also filter out the incorrect samples to reduce the spread of the misinformation. To show the importance of confidence estimation component, we conduct ablation study on multiplicative embedding methods, including DistMult, ComplEx, Analogy and SimplE. We add the proposed confidence estimation component into all the four multiplicative embedding methods.

\begin{figure}[hbt]
\vspace{-0.1in}
\subfloat[Recall]{
\vspace{-0.1in}
\begin{minipage}{.21\textwidth}
\includegraphics[height=1in]{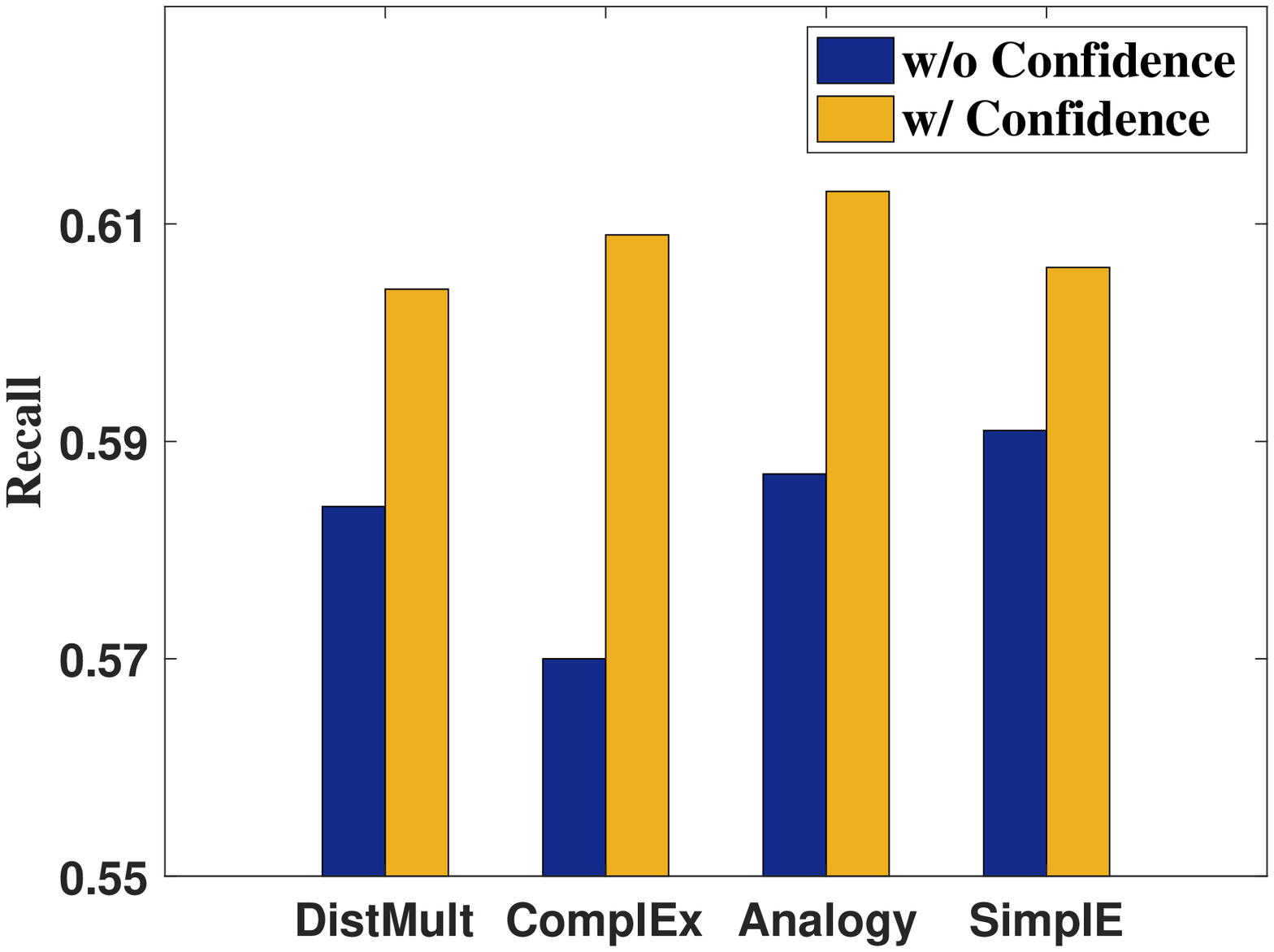}
\label{subfig:confidence_recall}
\end{minipage}}
\vspace{-0.1in}
\subfloat[Filter Mean Rank]{
\begin{minipage}{.23\textwidth}
\includegraphics[height=1in]{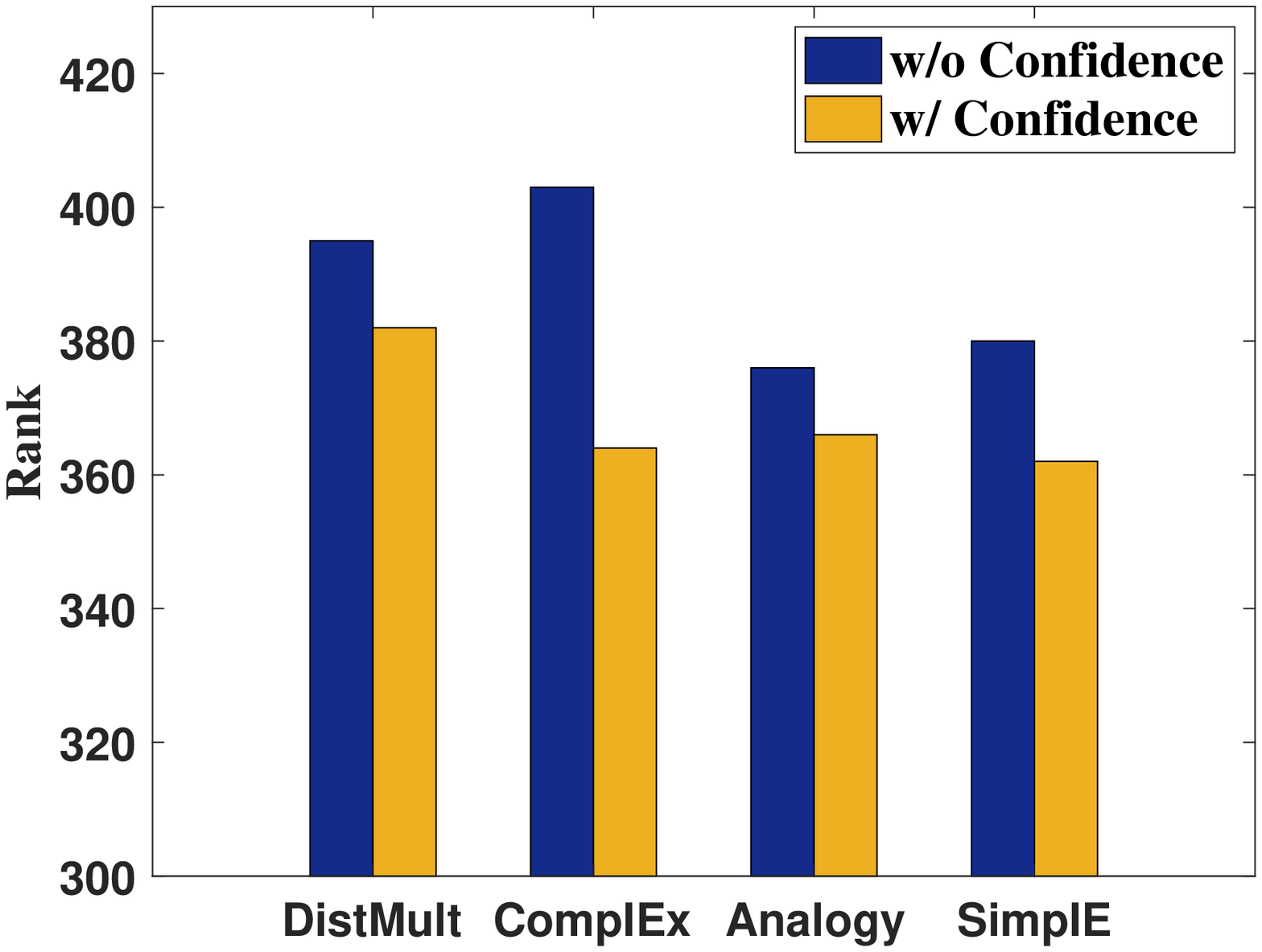}
\label{subfig:confidence_rank}
\end{minipage}}

 \caption{The performance comparison of methods w/ and w/o confidence estimation.}\label{fig:confidence}
  \vspace{-0.1in}
\end{figure}

Fig.~\ref{fig:confidence} shows the results in terms of \textit{Recall} and \textit{Filter Mean Rank}. In Fig.~\ref{fig:confidence}, ``w/ confidence'' means that we add confidence estimation component to the corresponding approaches, and ``w/o confidence'' denotes the original approaches. Take Fig.~\ref{subfig:confidence_recall} as an example, the \textit{Recall} of knowledge graph embedding methods ``w/ confidence'' is greater than that of the corresponding original approaches.  We can conclude that incorporating confidence estimation component is essential and effective for the task of error detection.


\textbf{Discussion.} The performance boost of the proposed framework mainly comes from the incorporation of the external KG, the confidence estimation and cross-KG negative sampling procedure. Take DistMult as an example,  the \textit{Recall} value of DistMult on NELL-314 is around 0.529. After incorporating the external KG via entity alignment and joint training, the \textit{Recall} value reaches 0.584. Adding confidence estimation component can improve the \textit{Recall} value to 0.604. The cross-negative sampling method can further improve the \textit{Recall} value from 0.604 to 0.641. 
We can observe that \emph{joint training},  \emph{cross-negative sampling}, and \emph{confidence estimation} all contribute to the performance boost.

\vspace{-0.05in}
\subsection{Hyperparameter Analysis}
\label{experiment:second_loss}

\textbf{Weight $\lambda$.}  We train the proposed framework using different weights $\lambda$ on the KG YAGO$_e$ and NELL-314. Fig.~\ref{fig:lambda} shows the \textit{Recall} changes of the proposed models with respect to different $\lambda$'s. When $\lambda$ is 0, the cross-KG negative sampling component is then removed from the proposed framework. We can observe that, without the cross-KG negative sampling methods, the proposed approaches cannot effectively take advantage of the information from the external KG. Thus, the recall values are lower compared with the corresponding proposed approaches with negative sampling component. After incorporating cross-KG negative sampling component, the proposed approaches achieve significant improvement in terms of \textit{Recall} values. Take \textsf{CrossVal}$_{Dist}$ as an example, when $\lambda$ is  $1$, the performance of the model achieves around 10\% improvement compared with the model without cross-KG sampling component. 
As observed from Fig.~\ref{fig:lambda}, the performance of the four approaches stay stable when $\lambda$ is larger than 0.1. This shows that the proposed framework is not very sensitive to the $\lambda$ value.


\textbf{Threshold $\theta$.} This threshold is used in the confidence estimation component to filter out samples with low confidence values. We train \textsf{CrossVal}$_{Dist}$ with the threshold values $0$, $0.3$, $0.5$ and $0.7$, and show the performance with respect to different $\theta$ values in Table~\ref{tab: threshold}. When $\theta$ is 0, \textsf{CrossVal}$_{Dist}$ can achieve the highest \textit{Recall} value among all the model variants but the lowest value on \textit{Precision@K}. As $\theta$ increases, the model can achieve higher value on \textit{Precision@K} but lower \textit{Recall} value. This may be because the model filters out more correct samples as the threshold value increases.

\begin{table}[htb]

\centering
\vspace{-0.1in}
\caption{The performance w.r.t.  different threshold $\theta$ values.} 
\vspace{-0.1in}
\label{tab: threshold}
\resizebox{0.7\linewidth}{!}{
\begin{tabular}{c|c|c|c|c|c|c}
\toprule

\multirow{2}{*}{$\theta$} & \multirow{2}{*}{Recall}&
\multicolumn{2}{c|}{Mean Rank}
&
\multicolumn{3}{c}{Precision@K} \\

\cline{3-7}
&&Filter&Raw & 
100  & 200 & 500\\
\midrule
\midrule

0&  \textbf{0.649} & 330 & 828 & 0.720 & 0.735 & 0.670   \\
0.3&0.643 & 330 & 828 & 0.770 & 0.755 & 0.668\\
0.5& 0.643 & \textbf{310}&\textbf{808}&\textbf{0.780}&\textbf{0.765}&\textbf{0.704}  \\
0.7& 0.641 & 320 & 819 & 0.850 & 0.745 & 0.676\\
\bottomrule
\end{tabular}
\vspace{-0.1in}
}

\vspace{-0.1in}
\end{table}

\textbf{Number of negative samples.} The number of negative samples is also an important hyperparameter for knowledge graph embedding model training. We train \textsf{CrossVal}$_{Dist}$ with different numbers of negative samples (i.e., 1, 5 and 10). The results are shown in Table~\ref{tab: neg_ent}. We can observe that adding more negative samples can help improve the \textit{Precision@K} values especially \textit{Precision@100} and \textit{Precision@200}. When the number of negative samples is set to be 10, the \textit{Recall} value decreases compared with the \textsf{CrossVal}$_{Dist}$ using 5 negative samples.

\begin{table}[htb]
\centering

\caption{The performance comparison w.r.t different number of negative samples.} 

\label{tab: neg_ent}
\resizebox{\linewidth}{!}{
\begin{tabular}{c|c|c|c|c|c|c}
\toprule

\multirow{2}{*}{\# of Negative Samples} & \multirow{2}{*}{Recall}&
\multicolumn{2}{c|}{Mean Rank}
&
\multicolumn{3}{c}{Precision@K} \\

\cline{3-7}
&&Filter&Raw & 
100  & 200 & 500\\
\midrule
\midrule

1&  0.641 & 305 & 803 & 0.780 & 0.720 & 0.704   \\
5 & \textbf{0.662}& \textbf{297} & \textbf{795} & 0.820 & 0.745 & \textbf{0.708}\\
10 & 0.643 & 310&808&\textbf{0.840}&\textbf{0.765}&0.704 \\
\bottomrule
\end{tabular}
}

\vspace{-0.1in}
\end{table}

\subsection{Analysis of the External KG}
 We conduct experiments to show the performance changes of the proposed framework when varying the size of the external KG and varying the number of overlapping entities between the external and target KGs.
 We sample different numbers of triplets  and sample different numbers of overlapping entities from the external KG for model training and report \textit{Recall} values of the proposed model \textsf{CrossVal}$_{Dist}$ on the real test dataset of NELL. 
 
  \textbf{Varying size of the external KG.} To clearly show the difference brought by the external KG, we use the \textit{Recall} value obtained from the baseline method \textsf{DistMult} that is trained on NELL-314 as the starting point. The triplet size of the incorporated external KG varies from 10,000 to 150,000. As can be seen in Fig.~\ref{fig:external_data}, only using 10k external triplets, the proposed model even achieves improvement around 10\% on \textit{Recall} compared with the baseline method \textsf{DistMult}. This demonstrates that the proposed framework is effective for KG validation even with a small external KG. 

\textbf{Varying size of the overlapping entities.} Since overlapping entities play an important role in bridging two KGs, we vary the size of overlapping entities (30\%, 50\%, 70\% and 100\%) to explore its impacts and report the performance in Table~\ref{tab: overlap_entity}. We can observe that the performance is improved as the size of overlapping entities increases. Moreover, 50\% of overlapping entities can help model achieve a \textit{Recall} value of 0.613. 

\begin{table}[htb]

\centering
\caption{The performance w.r.t.  different percentages of the overlapping entities between NELL-314 and YAGO$_e$.} 
\vspace{-0.1in}
\label{tab: overlap_entity}
\resizebox{\linewidth}{!}{
\begin{tabular}{c|c|c|c|c|c|c}
\toprule

\multirow{2}{*}{Percentage of Overlapping Entities} & \multirow{2}{*}{Recall}&
\multicolumn{2}{c|}{Mean Rank}
&
\multicolumn{3}{c}{Precision@K} \\

\cline{3-7}
&&Filter&Raw & 
100  & 200 & 500\\
\midrule
\midrule

30\%&  0.513 & 493 & 992 & 0.660 & 0.635 & 0.552   \\
50\%& 0.613 & 358 & 856 & 0.740 & 0.685 & 0.660\\
70\%& 0.633 & 324&823&0.760&0.695&0.678 \\
100\%& \textbf{0.641} & \textbf{305} & \textbf{803} & \textbf{0.780} & \textbf{0.720} & \textbf{0.704} \\
\bottomrule
\end{tabular}
}

\vspace{-0.1in}
\end{table}

\vspace{-0.1in}
\subsection{Efficiency Evaluation}
\label{sec:efficiency}

 The time complexity of the proposed framework is linear with respect to the number of the triplets. To clearly demonstrate this claim, we sample different numbers of triplets from the dataset to show the validation time of the proposed framework on a single GPU server. Take the proposed approach \textsf{CrossVal}$_{Dist}$ as an example. As shown in Fig.~\ref{fig:validation_time}, the proposed approach \textsf{CrossVal}$_{Dist}$ has linear time complexity with the number of triplets and validates 100k triplets  with less than 0.4 seconds. Based on the observations from Fig.~\ref{fig:validation_time}, we can  conclude that the proposed framework is efficient for large-scale KG validation.

\vspace{-0.1in}
\section{CONCLUSIONS}
\label{sec:con}
In this paper, we proposed to investigate the important problem of knowledge graph validation, which aims to detect errors by ranking all triplets in the knowledge graph according to their possibilities of being true. To tackle this problem, we proposed a novel framework that can leverage external knowledge graph and learn confidence aware knowledge graph embedding for error detection. 
Under the proposed framework, existing approaches have shown significantly improved performance for knowledge graph validation task on  datasets collected from both general and medical domains.  Furthermore, we analyze the role of each component of proposed framework, explore the effects of hyperparameters, investigate the performance with different amount of external information and show the efficiency of the proposed framework. This was demonstrated in a series of experiments conducted on the real datasets.
\vspace{-0.2in}
\section*{Acknowledgment}
The authors would like to thank the anonymous referees for their valuable comments and helpful suggestions. This work is supported in part by the US National Science Foundation under grants NSF-IIS 1747614 and NSF-IIS 1553411. Any opinions, findings, and conclusions or recommendations expressed in this material are those of the author(s) and do not necessarily reflect the views of the National Science Foundation.

\bibliographystyle{ACM-Reference-Format}
\bibliography{acmart} 

\end{document}